# Interpretation and Generalization of Score Matching

Siwei Lyu

Computer Science Department
University at Albany, SUNY
Albany, NY 12222, USA
lsw@cs.albany.edu

## Abstract

Score matching is a recently developed parameter learning method that is particularly effective to complicated high dimensional density models with intractable partition functions. In this paper, we study two issues that have not been completely resolved for score matching. First, we provide a formal link between maximum likelihood and score matching. Our analysis shows that score matching finds model parameters that are more robust with noisy training data. Second, we develop a generalization of score matching. Based on this generalization, we further demonstrate an extension of score matching to models of discrete data.

## 1 Introduction

Parameter estimation is an important task in machine learning and statistics. For statistical models of high dimensional data (e.g., Markov random fields [Win03]), the most commonly used method, *maximum likelihood*, may suffer from the intractable computation of the normalizing partition functions. Because of this, there exist several alternative learning schemes, notable examples include maximum pseudo-likelihood estimation [Bes74] and contrastive divergence [Hin02]. Recently, a new learning method, known as *score matching*, was introduced in [Hyv05], which has been shown to have several desirable properties. First, it leads to a consistent estimation, and the key optimization is deterministic. More important, the partition function in the parametric model has no effect in the optimization, and thus their computation is not necessary. It has been further shown that for certain exponential family models score matching also has close-form solutions [Hyv07b]. In practice, score matching has been applied to models for natural images [KLH09] and videos [CO09].

Notwithstanding these nice properties and successful applications, there are still two issues on score matching that have not been thoroughly studied. First, there lacks a clear relation between maximum likelihood and score matching, which will help to elucidate its excellent performance in practice. Second, the original score matching relies on properties of continuous data and differentiable models that do not hold for discrete data. It is thus desirable to develop a similar learning scheme that can apply to the latter case.

Correspondingly, this paper have two goals. First, we establish a formal link between maximum likelihood and score matching, by showing that the objective function in score matching is the derivative of that of maximum likelihood in the scale space of probability density functions with regards to the scale factor. This suggests an interesting interpretation of score matching that it seeks parameters that lead to models robust to noisy training data. Second, we provide a *generalized score matching* that extend the formulation of the original method while keeping its computational advantages. Based on this generalization, we demonstrate an extension of score matching for discrete data.

The rest of this paper is organized as following: in Section 2, we provide background of this work. Section 3 focuses on the formal relation between maximum likelihood and score matching. Section 4 introduces the generalized score matching and in Section 5, a specific instantiation of the generalized score matching is described for discrete data. Finally Section 6 concludes the paper with discussions.

## 2 Background

In statistical modeling, we are given $d$ dimensional data $\vec{x}$ with density $p(\vec{x})$, and our goal is to find a parametric probabilistic model $q_\theta(\vec{x})$, with $\theta$ being the model parameter, that best matches $p(\vec{x})$. To be more specific, we choose a divergence metric between probability density functions, and learning becomes finding parameter $\theta$ that minimizes such a divergence of $p(\vec{x})$ and $q_\theta(\vec{x})$. Ideally, the divergence metric should be non-negative, and becomes zero only when the two densities are equal almost everywhere. So if $p(\vec{x})$ is a member of the parametric density family of $q_\theta(\vec{x})$, the optimal parameter could be found by the learning procedure.

Based on this general view, the classical maximum likelihood learning, which finds $\theta$ to maximize the log likelihood function $\int_{\vec{x}} p(\vec{x}) \log q_\theta(\vec{x}) d\vec{x}$, can be equivalently understood as minimizing the Kulback-Leibler (KL) divergence between $p(\vec{x})$ and $q_\theta(\vec{x})$ [CT06]:

$$D_{KL}(p\|q_\theta) = \int_{\vec{x}} p(\vec{x}) \log \frac{p(\vec{x})}{q_\theta(\vec{x})} d\vec{x}. \quad (1)$$

This is because $D_{KL}(p\|q_\theta) = \int_{\vec{x}} p(\vec{x}) \log p(\vec{x}) d\vec{x} - \int_{\vec{x}} p(\vec{x}) \log q_\theta(\vec{x}) d\vec{x}$. The first term (the negative differential entropy of $p(\vec{x})$) is a constant with regards to $\theta$, and the second term is the log likelihood. Thus, minimizing KL divergence is equivalent to maximizing the log likelihood. However, in the case when we only have a large number of training data instead of the analytical form of $p(\vec{x})$, it is more convenient to work with the log likelihood function, as its dependency on $p(\vec{x})$ is through an expectation that can be approximated with averaging over the training data.

The maximum likelihood learning may become very difficult for models of high dimensional data. The problem originates from the requirement that $q_\theta(\vec{x})$ should be normalized to one, i.e. $\int_{\vec{x}} q_\theta(\vec{x}) d\vec{x} = 1$. For high dimensional data models such as the Markov random fields, they are typically defined using unnormalized components (e.g., clique potentials), and computing the normalizing constant, known as the *partition function*, can be intractable. As maximum likelihood learning relies on the direct computation of $q_\theta(\vec{x})$, this intractable partition function becomes a major computational bottleneck for learning high dimensional data models.

### 2.1 Score matching

Score matching is a different parameter learning methodology recently proposed in [Hyv05]. It cleverly obviates computing the partition function by using an alternative divergence metric of density functions, which we call the Fisher divergence. Formally, the Fisher divergence between $p(\vec{x})$ and $q_\theta(\vec{x})$ is defined as:

$$D_F(p\|q_\theta) = \int_{\vec{x}} p(\vec{x}) \left| \frac{\nabla_{\vec{x}} p(\vec{x})}{p(\vec{x})} - \frac{\nabla_{\vec{x}} q_\theta(\vec{x})}{q_\theta(\vec{x})} \right|^2 d\vec{x}, \quad (2)$$

where $\nabla_{\vec{x}}$ is the gradient operator with regards to $\vec{x}$. The learning procedure that finds $\theta$ to minimize $D_F(p\|q_\theta)$ is given the name score matching due to that $\frac{\nabla_{\vec{x}} p(\vec{x})}{p(\vec{x})}$ is known as the (Fisher) score function in statistics.

Just as the KL divergence is induced from the Shannon differential entropy, so does the Fisher divergence from the Fisher information[1],

$$J(p) = \int_{\vec{x}} p(\vec{x}) \left| \frac{\nabla p(\vec{x})}{p(\vec{x})} \right|^2 d\vec{x} = \int_{\vec{x}} p(\vec{x}) \left| \nabla \log p(\vec{x}) \right|^2 d\vec{x}. \quad (3)$$

Other similarities to the KL divergence include that the Fisher divergence is non-negative and is zero if and only if $p(\vec{x}) = q_\theta(\vec{x})$ (*a.e.*), yet it is not symmetric and does not form a distance metric.

It is not hard to see that in score matching, there is no need to use the partition function, in other words, it can work directly with un-normalized models. This is because in the score functions, $\frac{\nabla_{\vec{x}} p(\vec{x})}{p(\vec{x})}$ and $\frac{\nabla_{\vec{x}} q_\theta(\vec{x})}{q_\theta(\vec{x})}$, the partition functions appear in both the denominator and the numerator, which cancel out and thus have no effect to the Fisher divergence. Furthermore, as shown in [Hyv05], the squared distance of the model score function from the data score function, as measured by the Fisher divergence, can be computed as a simple expectation of functions of the un-normalized model. To better see this, first expand Eq.(2) as:

$$\int_{\vec{x}} p \left| \nabla_{\vec{x}} \log p \right|^2 + \int_{\vec{x}} p \left| \nabla_{\vec{x}} \log q_\theta \right|^2 - 2 \int_{\vec{x}} \nabla_{\vec{x}} p^T \frac{\nabla_{\vec{x}} q_\theta}{q_\theta}. \quad (4)$$

Assume that both $p(\vec{x})$ and $q_\theta(\vec{x})$ are smooth and fast decaying, such that their logarithms have growth at most polynomial at infinity. This implies that for both densities, we have

$$\lim_{x_i \to \pm\infty} p(\vec{x}) \frac{\partial^k}{\partial x_i^k} \log p(\vec{x}) = 0, \quad (5)$$

for any $i = 1, \cdots, d$ and certain non-negative integer $k$. We can transform, using integration by parts, the last term in Eq.(4), and use Eq.(5) to ensure that boundary values vanishes, to have:

$$\int_{\vec{x}} \nabla_{\vec{x}} p^T \frac{\nabla_{\vec{x}} q_\theta}{q_\theta} = -\int_{\vec{x}} p \nabla_{\vec{x}}^T \nabla_{\vec{x}} \log q_\theta = -\int_{\vec{x}} p \triangle_{\vec{x}} \log q_\theta,$$

where $\nabla_{\vec{x}}^T = \sum_{i=1}^d \frac{\partial}{\partial x_i}$ and $\triangle_{\vec{x}} = \nabla_{\vec{x}}^T \nabla_{\vec{x}} = \sum_{i=1}^d \frac{\partial^2}{\partial x_i^2}$ are the divergence and Laplacian operators, respectively. Subsequently, the Fisher divergence becomes

$$D_F(p\|q_\theta) = \int_{\vec{x}} p \left( \left| \nabla_{\vec{x}} \log p \right|^2 + \left| \nabla_{\vec{x}} \log q_\theta \right|^2 + 2 \triangle_{\vec{x}} \log q_\theta \right). \quad (6)$$

---

[1] Fisher information can be defined for any parameter in the density. The specific form given in Eq.(3) is for a location parameter [CT06].

As the first term is independent of the model parameter $\theta$, the dependency on $p(\vec{x})$ in searching for the optimal parameter reduces to the expectation. This is desirable when we only have observed data, where the expectation can be approximated with averaging over the training data set.

## 3 Score Matching and Maximum Likelihood

There is a striking similarity between the Fisher divergence and the KL divergence as in Eq.(1). If we rewrite the Fisher divergence, Eq.(2), as:

$$D_F(p\|q_\theta) = \int_{\vec{x}} p(\vec{x}) \left| \nabla_{\vec{x}} \log \frac{p(\vec{x})}{q_\theta(\vec{x})} \right|^2 d\vec{x},$$

their difference lie in that instead of using the likelihood ratio, the Fisher divergence computes the $l_2$ norm of the gradient of the likelihood ratio. This implies that there may be a deeper relation between them, and hence between score matching and maximum likelihood. This is indeed the case and is summarized in the following theorem.

**Theorem 1.** *Let $\vec{y} = \vec{x} + \sqrt{t}\vec{w}$, for $t \geq 0$ and $\vec{w}$ a zero-mean white Gaussian vector. Denote $\tilde{p}_t(\vec{y})$ and $\tilde{q}_t(\vec{y})$ as the densities of $\vec{y}$ when $\vec{x}$ has distribution $p(\vec{x})$ or $q(\vec{x})$, respectively. Assume that $\tilde{p}_t(\vec{y})$ and $\tilde{q}_t(\vec{y})$ are smooth and fast decaying, such that their logarithms has growth at most polynomial at infinity. We have*

$$\frac{d}{dt} D_{KL}(\tilde{p}_t(\vec{y})\|\tilde{q}_t(\vec{y})) = -\frac{1}{2} D_F(\tilde{p}_t(\vec{y})\|\tilde{q}_t(\vec{y})). \quad (7)$$

As $\tilde{p}_0(\vec{y}) = p(\vec{x})$ and $\tilde{q}_0(\vec{y}) = q(\vec{x})$, we further have $\frac{d}{dt} D_{KL}(\tilde{p}_t(\vec{y})\|\tilde{q}_t(\vec{y}))\big|_{t=0} = -\frac{1}{2} D_F(p(\vec{x})\|q(\vec{x}))$.

To prove Theorem 1, we need the following two lemmas, whose proofs are given in the Appendix.

**Lemma 1.** *For any positive valued function $f(\vec{x})$ whose gradient $\nabla_{\vec{x}}$ and Laplacian $\triangle_{\vec{x}}$ are well defined, we have identity*

$$\frac{\triangle_{\vec{x}} f(\vec{x})}{f(\vec{x})} = \triangle_{\vec{x}} \log f(\vec{x}) + |\nabla_{\vec{x}} \log f(\vec{x})|^2. \quad (8)$$

**Lemma 2.** *[Heat kernel] For density $\tilde{p}_t(\vec{y})$ as defined in Theorem 1, the following identity holds*

$$\frac{d}{dt} \tilde{p}_t(\vec{y}) = \frac{1}{2} \triangle_{\vec{y}} \tilde{p}_t(\vec{y}). \quad (9)$$

*Proof.* [Theorem 1] For conciseness in notation, we drop references to variables $\vec{x}$ and $\vec{y}$ in the integration, the density functions, and the operators whenever this does not lead to ambiguity.

First, with Lemma 1, $D_F(\tilde{p}\|\tilde{q})$ (Eq.(6)) becomes:

$$\begin{aligned} D_F(\tilde{p}\|\tilde{q}) &= \int \tilde{p} \left( |\nabla \log \tilde{p}|^2 + |\nabla \log \tilde{q}|^2 + 2\triangle \log \tilde{q} \right) \\ &= \int \tilde{p} \left( |\nabla \log \tilde{p}|^2 + \frac{\triangle \tilde{q}}{\tilde{q}} + \triangle \log \tilde{q} \right). \quad (10) \end{aligned}$$

Next, expanding the left hand side of Eq.(7), we have:

$$\frac{d}{dt} D_{KL}(\tilde{p}\|\tilde{q}) = \int \frac{d}{dt} \tilde{p} \log \frac{\tilde{p}}{\tilde{q}} + \int \tilde{p} \frac{d}{dt} \log \tilde{p} - \int \tilde{p} \frac{d}{dt} \log \tilde{q}.$$

We can eliminate the second term by exchanging integration and differentiation of $t$:

$$\int \tilde{p} \frac{d}{dt} \log \tilde{p} = \int \tilde{p} \frac{\frac{d\tilde{p}}{dt}}{\tilde{p}} = \int \frac{d}{dt} \tilde{p} = \frac{d}{dt} \int \tilde{p} = 0.$$

As a result, there are three remaining terms in computing $\frac{d}{dt} D_{KL}(\tilde{p}\|\tilde{q})$, which we can further substitute using Lemma 2, as:

$$\begin{aligned} \frac{d}{dt} D_{KL}(\tilde{p}\|\tilde{q}) &= \int \frac{d}{dt} \tilde{p} \log \tilde{p} - \int \frac{d}{dt} \tilde{p} \log \tilde{q} - \int \tilde{p} \frac{d}{dt} \log \tilde{q} \\ &= \frac{1}{2} \left( \int \triangle \tilde{p} \log \tilde{p} - \int \triangle \tilde{p} \log \tilde{q} - \int \tilde{p} \frac{\triangle \tilde{q}}{\tilde{q}} \right) \quad (11) \end{aligned}$$

Using integration by parts, the first term in Eq.(11) is changed to:

$$\int \triangle \tilde{p} \log \tilde{p} = \sum_{i=1}^{d} \left. \frac{\partial \tilde{p}}{\partial y_i} \log \tilde{p}(\vec{y}) \right|_{y_i=-\infty}^{y_i=\infty} - \int \nabla \tilde{p}^T \nabla \log \tilde{p}.$$

The limits in the first term becomes zero given the smoothness and fast decay properties of $\tilde{p}(\vec{y})$. The remaining term can be further simplified as:

$$\int \nabla \tilde{p}^T \nabla \log \tilde{p} = \int \tilde{p} \frac{(\nabla \tilde{p})^T}{\tilde{p}} \nabla \log \tilde{p} = \int \tilde{p} |\nabla \log \tilde{p}|^2.$$

The second term in Eq.(11) can be manipulated similarly, by first using integration by parts to get:

$$\int \triangle \tilde{p} \log \tilde{q} = \sum_{i=1}^{d} \left. \frac{\partial \tilde{p}}{\partial y_i} \log \tilde{q} \right|_{y_i=-\infty}^{y_i=\infty} - \int \nabla \tilde{p}^T \nabla \log \tilde{q}.$$

Applying integration by parts again to $\int \nabla \tilde{p}^T \nabla \log \tilde{q}$, we have

$$\int \nabla \tilde{p}^T \nabla \log \tilde{q} = \sum_{i=1}^{d} \left. \tilde{p} \frac{\partial \log \tilde{q}}{\partial y_i} \right|_{y_i=-\infty}^{y_i=\infty} - \int \tilde{p} \triangle \log \tilde{q}.$$

The limits at the boundary values are all zero due to the smoothness and fast decay properties of $\tilde{p}(\vec{y})$. Now collecting all terms, we have $\int \triangle \tilde{p} \log \tilde{p} = -\int \tilde{p} |\nabla \log \tilde{p}|^2$ and $\int \triangle \tilde{p} \log \tilde{q} = \int \tilde{p} \triangle \log \tilde{q}$, Eq.(11) becomes

$$\frac{d}{dt} D_{KL}(\tilde{p}\|\tilde{q}) = -\frac{1}{2} \int \tilde{p} \left( |\nabla \log \tilde{p}|^2 + \triangle \log \tilde{q} + \frac{\triangle \tilde{q}}{\tilde{q}} \right).$$

Combining with Eq. (10), this proves the result. □

Theorem 1 reveals some intriguing aspects of the relation between score matching and maximum likelihood by setting up a formal relation between the Fisher divergence and the KL divergence.

1. The effect of adding white Gaussian noise, $\sqrt{t}\vec{w}$, relates the density of $\vec{x}$ and $\vec{y}$ by

$$\tilde{p}_t(\vec{y}) = \int_{\vec{x}} \frac{1}{(2\pi t)^{d/2}} \exp\left(-\frac{|\vec{y}-\vec{x}|^2}{2t}\right) p(\vec{x})d\vec{x},$$

i.e., $\tilde{p}_t(\vec{y})$ is the convolution of $p(\vec{x})$ and a white Gaussian density of zero mean and variance level $t$. It is known that this process forms a scale space [Lin94] over probability densities, which composes Gaussian smoothed density functions of different scale factor $t$. With large scale factors, small local structures in the density function are smoothed. So if parameter in $q_\theta$ to match $p$ is sought in the scale space, it can put emphasis on large scale structures that survive the smoothing operation, and at the same time, spurious structures caused by the sampling effects of the training data are discounted. Indeed, this methodology has been adopted in clustering and non-parametric density estimation, e.g., [LZX00].

2. Theorem 1 elucidates that the Fisher divergence between two densities for scale factor $t$ equals to the derivative of their KL divergence with regards to the scale factor at the value of $t$. As the Fisher divergence between two densities are non-negative, this implies that the KL divergence between two densities never increases as the scale factor increases (or equivalently, the signal to noise ratio decreases). This is easy to understand, as the stronger noise is added, different signal sources get closer to the distribution of the noise and become more similar.

3. While maximum likelihood aims to minimize the KL divergence directly, according to Theorem 1, score matching seeks to eliminate its derivative in the scale space at $t = 0$. In other words, score matching looks for *stability*, where the optimal parameter $\theta$ leads to least changes in the KL divergence between the two models when a small amount of noise is present in the training data, while maximum likelihood pursues *extremity* of the KL divergence. It is known that maximum likelihood estimation is sensitive to noisy training data, which may give rise to many false extreme values, yet score matching may be more robust to small perturbation in training data. On the other hand, due to this fundamental difference, score matching and maximum likelihood can lead to very different solutions for the same parametric density family and training data, the only known exception of which is when $q_\theta$ is Gaussian [Hyv05].

4. There have been other interpretations of score matching based on data corrupted by additive Gaussian noise. In [Hyv08], it was shown that score matching is an approximation of the optimal parameter estimation when using the model as a prior in the inference of noise-free signal, and as the noise goes to infinitesimally small. In [RS07], score matching was interpreted as searching parameters of $q_\theta$ so that when a Bayes least square estimator is constructed based on it, the overall mean square error with the optimal estimator based on $p(\vec{x})$ is minimal (see Section 4 for more details). However, neither of these provide a direct relation between score matching and maximum likelihood.

5. Finally, as a special case of Eq.(7), when $\tilde{q}_t$ is set to a uniform distribution over the support of $\tilde{p}_t$ so that $\frac{d}{dt}D_{KL}(\tilde{p}\|\tilde{q}) = \frac{d}{dt}H(\tilde{p}_t)$ and $D_F(\tilde{p}\|\tilde{q}) = J(\tilde{p}_t)$, where $H(p) = -\int_{\vec{x}} p(\vec{x})\log p(\vec{x})d\vec{x}$ is the (Shannon) differential entropy and $J(\tilde{p}_t)$ is the Fisher information as defined in Eq.(3), we have

$$\frac{d}{dt}H(\tilde{p}_t) = \frac{1}{2}J(\tilde{p}_t).$$

This is a well known result in the information theory as the *de Bruijn's identity*, which reveals a remarkable geometric relation between the differential entropy and the Fisher information: the former is related to the volume of the typical set of $\tilde{p}_t$, the latter is related to its surface area [CT06].

## 4 Generalized Score Matching

The score matching learning can be generalized to a more flexible parametric learning methodology. Starting with the definition of the Fisher divergence, Eq.(2), the main idea is to replace the gradient, which is a linear operator (functional) on density functions, with a general linear operator $\mathcal{L}$, as:

$$D_{\mathcal{L}}(p\|q_\theta) = \int_{\vec{x}} p \left|\frac{\mathcal{L}p(\vec{x})}{p(\vec{x})} - \frac{\mathcal{L}q_\theta(\vec{x})}{q_\theta(\vec{x})}\right|^2 d\vec{x}. \qquad (12)$$

If $\vec{x}$ has discrete components, integration is substituted with summations. We term $D_{\mathcal{L}}$ the *generalized Fisher divergence*, and $\frac{\mathcal{L}p(\vec{x})}{p(\vec{x})}$ the *generalized score function*. Correspondingly, parametric learning using $D_{\mathcal{L}}$ is called the *generalized score matching*. It is easy to see that $D_{\mathcal{L}}$ is non-negative, and is zero when the two densities equal almost everywhere.

The generalized Fisher divergence keeps several important computational advantages of the original Fisher divergence. First, as an linear operator does not affect the normalizing partition function, it is canceled out in the generalized score function, and hence has no effect in the subsequent computation. Second, the generalized Fisher divergence can also be transformed to a form as an expectation of functions of the unnormalized model. To see this, we need the following definition of $\mathcal{L}$'s adjoint.

**Definition 1.** *Denote $\mathcal{F}^1$ and $\mathcal{F}^D$ as the space of all scalar-valued and D-variate functions for $\vec{x}$, respectively. $\mathcal{L} : \mathcal{F}^1 \mapsto \mathcal{F}^D$ is an linear operator. Further, assume that both $\mathcal{L}f(\vec{x})$ and $g(\vec{x})$ are square integrable, i.e., $\int_{\vec{x}} |\mathcal{L}f(\vec{x})|^2 d\vec{x} < \infty$ and $\int_{\vec{x}} |g(\vec{x})|^2 d\vec{x} < \infty$. The adjoint of $\mathcal{L}$, $\mathcal{L}^+ : \mathcal{F}^D \mapsto \mathcal{F}^1$, is a linear operator satisfying that $\forall f \in \mathcal{F}^1$ and $g \in \mathcal{F}^D$,*

$$\int_{\vec{x}} (\mathcal{L}f(\vec{x}))^T g(\vec{x}) d\vec{x} = \int_{\vec{x}} f(\vec{x})(\mathcal{L}^+ g(\vec{x})) d\vec{x}.$$

Next, expand $D_\mathcal{L}(p\|q_\theta)$

$$D_\mathcal{L}(p\|q_\theta) = \int_{\vec{x}} p \left[ \left|\frac{\mathcal{L}p}{p}\right|^2 + \left|\frac{\mathcal{L}q_\theta}{q_\theta}\right|^2 - 2\left(\frac{\mathcal{L}p}{p}\right)^T \left(\frac{\mathcal{L}q_\theta}{q_\theta}\right) \right].$$

Using the definition of adjoint operator, we rewrite:

$$\int_{\vec{x}} (\mathcal{L}p)^T \frac{\mathcal{L}q_\theta}{q_\theta} = \int p \mathcal{L}^+ \left(\frac{\mathcal{L}q_\theta}{q_\theta}\right),$$

which results in

$$D_\mathcal{L}(p\|q_\theta) = \int_{\vec{x}} p \left[ \left|\frac{\mathcal{L}p}{p}\right|^2 + \left|\frac{\mathcal{L}q_\theta}{q_\theta}\right|^2 - 2\mathcal{L}^+\left(\frac{\mathcal{L}q_\theta}{q_\theta}\right) \right] \quad (13)$$

As the first term is a constant with regards to the model parameters, we only need $p$ for computing expectations. Thus, the generalized score matching can also be computed from training data as in the case of the score matching.

Difference choices of $\mathcal{L}$ lead to different instantiations of generalized score matching. Though in principle we can use any linear operator, for parameter learning, we need operators leading to score functions that do not "lose" information about the original density. This is formalized in the following definition.

**Definition 2.** *A linear operator $\mathcal{L}$ is* complete *if for two densities $p(\vec{x})$ and $q(\vec{x})$, $\frac{\mathcal{L}p(\vec{x})}{p(\vec{x})} = \frac{\mathcal{L}q(\vec{x})}{q(\vec{x})}$ (almost everywhere) $\Rightarrow p(\vec{x}) = q(\vec{x})$ (almost everywhere). Otherwise, it is* incomplete.

An extreme example of incomplete operator is $\mathcal{L}f(\vec{x}) = 0$ for any $f(\vec{x})$, with which for any two densities, we have $D_\mathcal{L}(p\|q) = 0$. However, it is obvious that this operator is no use for parameter learning. In the following, we show some simple choices for $\mathcal{L}$.

**Gradient.** If we choose $\mathcal{L}$ to be the gradient operator, $\nabla$, $D_\mathcal{L}$ reduces to the original Fisher divergence, and the corresponding generalized score matching becomes the original score matching. As a confirmation, note that the adjoint of gradient is the negative divergence, i.e., $-\nabla^T = \sum_{i=1}^d \frac{\partial}{\partial x_i}$, and $-\nabla^T \nabla = -\triangle$. Using these results, it is easy to see that Eq.(13) reduces to Eq.(6) in this case.

It is also easy to establish that the gradient operator is complete. If we have $\frac{\nabla p(\vec{x})}{p(\vec{x})} = \frac{\nabla q(\vec{x})}{q(\vec{x})}$[2], or equivalently, $\nabla \log p(\vec{x}) = \nabla \log q(\vec{x})$, this leads to $\nabla \log \frac{p(\vec{x})}{q(\vec{x})} = 0$, or $\frac{p(\vec{x})}{q(\vec{x})} = c$, where $c$ is a constant for all $\vec{x}$. Because both $p(\vec{x})$ and $q(\vec{x})$ are density functions, it must be that $c = 1$, which then implies $p(\vec{x}) = q(\vec{x})$.

**Marginalization.** Another choice for $\mathcal{L}$ is what we call the marginalization operator, $\mathcal{M} : \mathcal{F}^1 \mapsto \mathcal{F}^d$, which is defined as

$$\mathcal{M}f(\vec{x}) = \begin{pmatrix} \vdots \\ \mathcal{M}_i f(\vec{x}) \\ \vdots \end{pmatrix} = \begin{pmatrix} \vdots \\ \int_{x_i} f(\vec{x}) dx_i \\ \vdots \end{pmatrix}, \quad (14)$$

for any $f \in \mathcal{F}^1$. Integration is to be replaced with summation when $\vec{x}$ has discrete components. If $f(\vec{x}) = p(\vec{x})$ is a probability density over $\vec{x}$, $\mathcal{M}_i p(\vec{x})$ is the marginal density of $\vec{x}^{\backslash i}$ induced from $p(\vec{x})$, where $\vec{x}^{\backslash i}$ denotes the vector formed by dropping $x_i$ from $\vec{x}$ (hence the name of $\mathcal{M}$ as the marginalization operator). Consequently, we have

$$\frac{\mathcal{M}_i p(\vec{x})}{p(\vec{x})} = \frac{p(\vec{x}^{\backslash i})}{p(\vec{x})} = \frac{1}{p(x_i|\vec{x}^{\backslash i})}, \quad (15)$$

in other words, each component of $\mathcal{M}p(\vec{x})$ is the reciprocal of conditional density $p(x_i|\vec{x}^{\backslash i})$ induced from $p(\vec{x})$. Without losing generality, we assume here that $p(x_i|\vec{x}^{\backslash i}) \neq 0$. Such conditional densities are known as the singleton conditionals. Therefore, minimizing the generalized Fisher divergence between $p(\vec{x})$ and $q(\vec{x})$ under the marginalization operator is equivalent to match their corresponding singletons under the induced divergence. The marginalization operator is complete because of the following well known result in statistics.

**Lemma 3** (Brook's Lemma [Bro64])**.** *The joint density of random variables $(x_1, \cdots, x_d)$ are completely determined by the ensemble of singleton conditional densities, $p(x_i|\vec{x}^{\backslash i})$, $\forall i$.*

For completeness, we include a proof of this lemma in the Appendix.

In using the marginalization for the case where we only have training data instead of analytical form for $p$, we need Eq.(13) and the adjoint of $\mathcal{M}$, which is given by the following lemma.

---

[2]For simplicity, we drop the qualifier (*almost everywhere*) and assume $p(\vec{x})$ and $q(\vec{x})$ have the same support, yet it is not hard to enforce this condition in the subsequent description.

**Lemma 4.** *For $g(\vec{x}) \in \mathcal{F}^d$, and denote $(g(\vec{x}))_i = g_i(\vec{x})$, and $f(\vec{x}) \in \mathcal{F}^1$, it holds that*

$$\int_{\vec{x}} \mathcal{M}(f(\vec{x}))^T g(\vec{x}) d\vec{x} = \int_{\vec{x}} f(\vec{x}) \sum_{i=1}^{d} \mathcal{M}_i g_i(\vec{x}) d\vec{x},$$

*in other words, $\mathcal{M}^+ = \sum_{i=1}^{d} \mathcal{M}_i$.*

The proof is given in the Appendix.

**Posterior mean.** In [RS07], the generalized score function, $\mathcal{L}p/p$, was given a very different statistical interpretation. Assume $p$ is the density for variable $\vec{y}$, which depends on a latent variable $\vec{x}$. The mean for the posterior distribution, $p_{X|Y}(\vec{x}|\vec{y})$, defined as $E(\vec{x}|\vec{y}) = \int_{\vec{x}} \vec{x} p_{X|Y}(\vec{x}|\vec{y}) d\vec{x}$, is shown to take the form $\mathcal{L}p(\vec{y})/p(\vec{y})$, where $\mathcal{L}$ is determined from the conditional density $p_{Y|X}(\vec{y}|\vec{x})$. As $E(\vec{x}|\vec{y})$ is the optimal estimator of $\vec{x}$ given $\vec{y}$ that minimizes the mean square errors with $\vec{x}$, optimizing the resulting generalized Fisher divergence is equivalent to find the optimal density $q$ such that when used as a model for $\vec{y}$, it achieves the best performance in the inference of $\vec{x}$. Especially, when $\vec{y}$ is obtained by adding noise of some known density to $\vec{x}$, $\mathcal{L}$ and its adjoint may have simple close form solutions (e.g., the additive Gaussian noise case corresponds to the Fisher divergence and original score matching). However, not all complete linear operators suitable for $\mathcal{L}$ affords such an interpretation, such as the marginalization operator $\mathcal{M}$. On the other hand, it is hard to check the completeness of operators originated from the posterior means in general.

## 5 Generalized Score Matching for Discrete Data

There are two important restrictions in the original score matching method, being that $\vec{x}$ has to be continuous-valued and the densities have to be differentiable in the space of $\mathcal{R}^d$. Due to these restrictions, one cannot directly apply score matching to discrete data as $\nabla \log p(\vec{x})$ is not well defined in such cases. In this section, we show that using the marginalization operator $\mathcal{M}$ (Eq.(14)) with the generalized score matching leads to a natural extension of score matching to discrete data.

Consider discrete vectors $\vec{x} \in \{c_1, \cdots, c_m\}^d$ with density $p(\vec{x})$. Correspondingly, the integration in Eq.(12) is replaced with summation. As in the continuous case, learning is to find the optimal parameter for $q_\theta(\vec{x})$ that minimizes its generalized Fisher divergence with $p(\vec{x})$, which is

$$D_\mathcal{M}(p\|q_\theta) = \sum_{\vec{x}} p(\vec{x}) \sum_{i=1}^{d} \left( \frac{\mathcal{M}_i p(\vec{x})}{p(\vec{x})} - \frac{\mathcal{M}_i q_\theta(\vec{x})}{q_\theta(\vec{x})} \right)^2.$$

Substituting with Eq.(15), and using dummy variable $\xi_i$ where we need to marginalize over the $i^{th}$ component of $\vec{x}$ in the inner integral, we have

$$D_\mathcal{M}(p\|q_\theta) = \sum_{\vec{x}} p(\vec{x}) \sum_{i=1}^{d} \sum_{\xi_i} \left( p(\xi_i|\vec{x}^{\backslash i}) - q_\theta(\xi_i|\vec{x}^{\backslash i}) \right)^2 \tag{16}$$

On the other hand, with Eq.(13) and the adjoint operator of $\mathcal{M}$ (Lemma 4), it is further expanded to:

$$\sum_{\vec{x}} p(\vec{x}) \sum_{i=1}^{d} \left[ \left( \frac{\mathcal{M}_i p}{p} \right)^2 + \left( \frac{\mathcal{M}_i q_\theta}{q_\theta} \right)^2 - 2\mathcal{M}_i \left( \frac{\mathcal{M}_i q_\theta}{q_\theta} \right) \right].$$

Dropping the first term that is independent of $\theta$, we have:

$$\sum_{\vec{x}} p(\vec{x}) \sum_{i=1}^{d} \left[ \left( \frac{\mathcal{M}_i q_\theta}{q_\theta} \right)^2 - 2\mathcal{M}_i \left( \frac{\mathcal{M}_i q_\theta}{q_\theta} \right) \right]$$

$$= \sum_{\vec{x}} p(\vec{x}) \sum_{i=1}^{d} \sum_{\xi_i} \left[ \frac{1}{q_\theta^2(\xi_i|\vec{x}^{\backslash i})} - \frac{2}{q_\theta(\xi_i|\vec{x}^{\backslash i})} \right]$$

$$= \sum_{\vec{x}} p(\vec{x}) \sum_{i=1}^{d} \sum_{\xi_i} \frac{1 - 2q_\theta(\xi_i|\vec{x}^{\backslash i})}{q_\theta^2(\xi_i|\vec{x}^{\backslash i})}.$$

Further rearrangement leads to:

$$\sum_{\vec{x}} p(\vec{x}) \sum_{i=1}^{d} \sum_{\xi_i} \left( \frac{1 - 2q_\theta(\xi_i|\vec{x}^{\backslash i}) + q_\theta^2(\xi_i|\vec{x}^{\backslash i})}{q_\theta^2(\xi_i|\vec{x}^{\backslash i})} - 1 \right)$$

$$= \sum_{\vec{x}} p(\vec{x}) \sum_{i=1}^{d} \sum_{\xi_i} \frac{(1 - q_\theta(\xi_i|\vec{x}^{\backslash i}))^2}{q_\theta^2(\xi_i|\vec{x}^{\backslash i})} - md$$

$$= \sum_{\vec{x}} p(\vec{x}) \sum_{i=1}^{d} \sum_{\xi_i} \left( \frac{q_\theta(\sim \xi_i|\vec{x}^{\backslash i})}{q_\theta(\xi_i|\vec{x}^{\backslash i})} \right)^2 - md, \tag{17}$$

where we use $q_\theta(\sim \xi_i|\vec{x}^{\backslash i})$ to shorthand the conditional probability for the $i^{\text{th}}$ element of $\vec{x}$ not taking value $\xi_i$. As $\sum_{\xi_i} q_\theta(\xi_i|\vec{x}^{\backslash i}) = 1$, and $\sum_{\xi_i} \frac{(1-q_\theta(\xi_i|\vec{x}^{\backslash i}))^2}{q_\theta^2(\xi_i|\vec{x}^{\backslash i})}$ reaches its minimum when $q_\theta(\xi_i|\vec{x}^{\backslash i})$ approaches constant value, minimizing the generalized Fisher divergence has an overall effect of balancing the values of the singleton conditional densities.

### 5.1 Relation with Ratio Matching

We compare the aforementioned discrete extension of generalized score matching with another similar method, known as *ratio matching* [Hyv07a]. Originally, the ratio matching algorithm was described for binary data. Here we describe an extended version that can be applied to general discrete data types. First define a scalar function $\phi$, as $\phi(u) = \frac{1}{1+u}$, for $u \in$

$\mathcal{R}^+$. In ratio matching [Hyv07a], we find the optimal parameter $\theta$ that minimizes

$$\sum_{\vec{x}} p(\vec{x}) \sum_{i=1}^{d} \sum_{\xi_i} \left[ \phi\left(\frac{p(\xi_i, \vec{x}^{\backslash i})}{p(\sim\xi_i, \vec{x}^{\backslash i})}\right) - \phi\left(\frac{q_\theta(\xi_i, \vec{x}^{\backslash i})}{q_\theta(\sim\xi_i, \vec{x}^{\backslash i})}\right) \right]^2.$$

Here we use $q_\theta(\sim\xi_i, \vec{x}^{\backslash i})$ to denote the joint probability with $\vec{x}^{\backslash i}$ fixed and $x_i$ taking values other than $\xi_i$. This is slightly different from that used in [Hyv07a], as a result of merging identical terms and dropping irrelevant terms. Using the definition of $\phi$, we have $\phi\left(\frac{p(\xi_i, \vec{x}^{\backslash i})}{p(\sim\xi_i, \vec{x}^{\backslash i})}\right) = p(\sim\xi_i|\vec{x}^{\backslash i})$. The ratio matching objective function is:

$$\sum_{\vec{x}} p(\vec{x}) \sum_{i=1}^{d} \sum_{\xi_i} \left( p(\sim\xi_i|\vec{x}^{\backslash i}) - q_\theta(\sim\xi_i|\vec{x}^{\backslash i}) \right)^2$$
$$= \sum_{\vec{x}} p(\vec{x}) \sum_{i=1}^{d} \sum_{\xi_i} \left( p(\xi_i|\vec{x}^{\backslash i}) - q_\theta(\xi_i|\vec{x}^{\backslash i}) \right)^2.$$

Note the similarity of this function to that in our extension, Eq.(16). As shown in [Hyv07a], the ratio matching objective function can be further reduced to

$$\sum_{\vec{x}} p(\vec{x}) \sum_{i=1}^{d} \sum_{\xi_i} \left( \phi\left(\frac{q_\theta(\xi_i, \vec{x}^{\backslash i})}{q_\theta(\sim\xi_i, \vec{x}^{\backslash i})}\right) \right)^2$$
$$= \sum_{\vec{x}} p(\vec{x}) \sum_{i=1}^{d} \sum_{\xi_i} \left( 1 - q_\theta(\xi_i|\vec{x}^{\backslash i}) \right)^2 + \text{const.}$$

Note that the minimum of $\sum_{\xi_i} \left(1 - q_\theta(\xi_i|\vec{x}^{\backslash i})\right)^2$ is also reached when $q_\theta(\xi_i|\vec{x}^{\backslash i})$ is a constant, therefore at optimum, ratio matching and our extension agree with each other, and both of them are different from maximum pseudo-likelihood. On the other hand, note that the objective function in ratio matching is quite different from that in score matching [Hyv07a].

## 6 Conclusion

In this paper, we show two new results regarding the recently developed parameter learning method known as score matching. First, we establish a formal link between maximum likelihood and score matching, by showing the relation between the corresponding divergence functions. Specifically, we show that the Fisher divergence is the derivative of the KL divergence in a scale space with regards to the scale factor. This suggests that score matching searches for parameters that are stable with small noise perturbation in training data. Second, we provide a generalization of score matching by employing general linear operators in the Fisher divergence, and demonstrate a specific instantiation of the generalized score matching to discrete data to be a more natural extension of score matching to discrete data.

There are several directions that we hope to further explore in the future. First, by using other type of diffusion kernels, it may be possible to establish a similar relation between the maximum likelihood and the generalized score matching. Secondly, the generalized score matching provides more flexibility in applying the score matching methodology to different parameter estimation problems. Especially, it will be of great interest to study appropriate complete linear operators for specific high dimensional data models such as Markov random fields. Finally, we are currently working on applying the generalized score matching learning to practical problems such as bioinformatics and image modeling. We hope the work presented in this paper may deepen our understanding on score matching and help to extend its applications in machine learning and related fields.

## Appendix

*Proof.* [Lemma 1] For conciseness, we drop $\vec{x}$ from the gradient and Laplacian operator. Using the definition of Laplacian, we have

$$\triangle \log f(\vec{x}) = \frac{\nabla^T \nabla f(\vec{x})}{f(\vec{x})} - \left(\frac{\nabla f(\vec{x})}{f(\vec{x})}\right)^T \frac{\nabla f(\vec{x})}{f(\vec{x})}$$
$$= \frac{\triangle f(\vec{x})}{f(\vec{x})} - |\nabla \log f(\vec{x})|^2.$$

Rearranging terms proves the Lemma. $\square$

*Proof.* [Lemma 2] First, use the definition of $\tilde{p}_t(\vec{y})$,

$$\tilde{p}_t(\vec{y}) = \int_{\vec{x}} \frac{1}{(2\pi t)^{d/2}} \exp\left(-\frac{|\vec{y}-\vec{x}|^2}{2t}\right) p(\vec{x}),$$

and

$$\frac{d}{dt}\tilde{p}_t(\vec{y}) = \int_{\vec{x}} \frac{|\vec{y}-\vec{x}|^2}{2t^2} \frac{1}{(2\pi t)^{d/2}} \exp\left(-\frac{|\vec{y}-\vec{x}|^2}{2t}\right) p(\vec{x})$$
$$- \int_{\vec{x}} \frac{d}{2t} \frac{1}{(2\pi t)^{d/2}} \exp\left(-\frac{|\vec{y}-\vec{x}|^2}{2t}\right) p(\vec{x}). \quad (18)$$

On the other hand, note that

$$\nabla \tilde{p}_t(\vec{y}) = -\int_{\vec{x}} \frac{(\vec{y}-\vec{x})}{t} \frac{1}{(2\pi t)^{d/2}} \exp\left(-\frac{|\vec{y}-\vec{x}|^2}{2t}\right) p(\vec{x}).$$

Taking derivative again, we have

$$\triangle \tilde{p}_t(\vec{y}) = \int_{\vec{x}} \frac{|\vec{y}-\vec{x}|^2}{t^2} \frac{1}{(2\pi t)^{d/2}} \exp\left(-\frac{|\vec{y}-\vec{x}|^2}{2t}\right) p(\vec{x})$$
$$- \int_{\vec{x}} \frac{d}{t} \frac{1}{(2\pi t)^{d/2}} \exp\left(-\frac{|\vec{y}-\vec{x}|^2}{2t}\right) p(\vec{x}). \quad (19)$$

Combining (18) and (19) proves the lemma. $\square$

*Proof.* [Lemma 3] We prove this by showing that the ratio of the joint probability of two assignments, $(\xi_1, \cdots, \xi_n)$ and $(\tilde{\xi}_1, \cdots, \tilde{\xi}_n)$, of random variables $(x_1, \cdots, x_n)$, $p(\xi_1, \cdots, \xi_n)/p(\tilde{\xi}_1, \cdots, \tilde{\xi}_n)$, can be determined using only the singleton conditionals. As the joint density has to be normalized to one, this shows the uniqueness of the joint density given the singletons.

$$\frac{p(\xi_1, \cdots, \xi_n)}{p(\tilde{\xi}_1, \cdots, \tilde{\xi}_n)} = \frac{p(\xi_1, \xi_2, \cdots, \xi_n)}{p(\tilde{\xi}_1, \xi_2, \cdots, \xi_n)} \frac{p(\tilde{\xi}_1, \xi_2, \xi_3 \cdots, \xi_n)}{p(\tilde{\xi}_1, \tilde{\xi}_2, \xi_3 \cdots, \xi_n)}$$
$$\cdots \frac{p(\tilde{\xi}_1, \cdots, \xi_{n-1}, \xi_n)}{p(\tilde{\xi}_1, \cdots, \tilde{\xi}_{n-1}, \xi_n)} \frac{p(\tilde{\xi}_1, \cdots, \tilde{\xi}_{n-1}, \xi_n)}{p(\tilde{\xi}_1, \cdots, \tilde{\xi}_{n-1}, \tilde{\xi}_n)}$$
$$= \frac{p(\xi_1|\xi_2, \cdots, \xi_n)}{p(\tilde{\xi}_1|\xi_2, \cdots, \xi_n)} \frac{p(\xi_2|\tilde{\xi}_1, \xi_3 \cdots, \xi_n)}{p(\tilde{\xi}_2|\tilde{\xi}_1, \xi_3 \cdots, \xi_n)}$$
$$\cdots \frac{p(\xi_{n-1}|\tilde{\xi}_1, \cdots, \xi_n)}{p(\tilde{\xi}_{n-1}|\tilde{\xi}_1, \cdots, \xi_n)} \frac{p(\xi_n|\tilde{\xi}_1, \cdots, \tilde{\xi}_{n-1})}{p(\tilde{\xi}_n|\tilde{\xi}_1, \cdots, \tilde{\xi}_{n-1})}.$$

In the first step, we introduce terms that cancel out each other. The second step then cancels out common joint densities from each ratio, and result is completely determined by the singleton conditionals. □

*Proof.* [Lemma 4] We use the shorthand notation $f(\xi_i, \vec{x}^{\setminus i})$ for the otherwise longer notation $f(x_1, \cdots, x_{i-1}, \xi_i, x_{i+1}, \cdots, x_d)$ to emphasize on one variable.

$$\int_{\vec{x}} \mathcal{M}(f(\vec{x}))^T g(\vec{x}) d\vec{x} = \int_{\vec{x}} \sum_{i=1}^d \mathcal{M}_i f(\vec{x}) g_i(\vec{x}) d\vec{x}$$
$$= \int_{\vec{x}^{\setminus i}} \sum_{i=1}^d \int_{x_i} \int_{\xi_i} f(\xi_i, \vec{x}^{\setminus i}) g_i(x_i, \vec{x}^{\setminus i}) d\xi_i dx_i d\vec{x}^{\setminus i}$$

Define $\vec{x}' = (x_1, \cdots, x_{i-1}, \xi_i, x_{i+1}, \cdots, x_d)$, and switch the integration order for $x_i$ and $\xi_i$, from which the last step above can be rewritten as

$$\int_{\vec{x}'} f(\vec{x}') \sum_{i=1}^d \int_{\xi_i} g_i(\xi_i, \vec{x}^{\setminus i}) d\xi_i d\vec{x}',$$

which is equivalent to $\int_{\vec{x}'} f(\vec{x}') \sum_{i=1}^d \mathcal{M}_i g_i(\vec{x}') d\vec{x}'$. □